\def\year{2020}\relax
\documentclass[letterpaper]{article}
\usepackage{aaai20}  
\nocopyright
\usepackage{times}  
\usepackage{helvet} 
\usepackage{courier}  
\usepackage[hyphens]{url}  
\usepackage{graphicx} 
\usepackage{todonotes}
\usepackage{natbib}
\usepackage{amsmath}
\usepackage{graphicx}
\usepackage{tabularx}
\usepackage{soul}
\usepackage{todonotes}
\usepackage{tcolorbox}
\usepackage{color,soul}
\usepackage[section]{placeins}
\usepackage{epstopdf}
\usepackage[utf8]{inputenc}
\usepackage{multirow,tabularx}
\usepackage{devanagari}
\RequirePackage{colortbl}
\definecolor{airforceblue}{rgb}{0.36, 0.54, 0.66}
\definecolor{amaranth}{rgb}{0.9, 0.17, 0.31}
\definecolor{applegreen}{rgb}{0.55, 0.71, 0.0}
\definecolor{alizarin}{rgb}{0.82, 0.1, 0.26}
\definecolor{azure}{rgb}{0.0, 0.5, 1.0}
\definecolor{cadmiumgreen}{rgb}{0.0, 0.42, 0.24}

\usepackage{graphicx}  
\title{PHINC: A Parallel Hinglish Social Media Code-Mixed Corpus for Machine Translation}

\author{\Large \textbf{Vivek Srivastava, Mayank Singh}\\ 
Indian Institute of Technology Gandhinagar\\ 
Gujarat India 382355\\
vivek.srivastava@iitgn.ac.in, singh.mayank@iitgn.ac.in 
}

\begin{document}

\maketitle

\begin{abstract}
Code-mixing is the phenomenon of using more than one language in a sentence. It is a very frequently observed pattern of communication on social media platforms. Flexibility to use multiple languages in one text message might help to communicate efficiently with the target audience. But, it adds to the challenge of processing and understanding natural language to a much larger extent. This paper presents a parallel corpus of the 13,738 code-mixed English-Hindi sentences and their corresponding translation in English. The translations of sentences are done manually by the annotators. We are releasing the parallel corpus to facilitate future research opportunities in code-mixed machine translation. The annotated corpus is available at  \url{https://doi.org/10.5281/zenodo.3605597}.
\end{abstract}

\section{Introduction}

Code-mixing is the phenomenon of switching between two or more languages by the speaker in a single sentence of a text or speech. Code mixing is one of the most frequent styles of communication in multilingual communities, such as India. High spelling variations in the Romanized Hindi words (e.g., \textit{
mujhe, mjhe, mujhee, etc.}, are some variants for the Hindi counterpart of the English word \textit{me}.) presents the challenge to effective machine translation task. We frquently observe code-mixing on social media platforms such as Twitter, Facebook, etc. in contrast to the formal literature sources such as books, poems, and newspapers. We, therefore, use several social media platforms such as  Twitter, Facebook, etc. as the source of data for our purpose.

With the ever-increasing volume of user engagement on social media platforms, there is an upraise in the interest to study and build systems that support code-mixing of multiple resource-constraint Indian languages. \cite{barman-etal-2014-code} discuss the language identification task for the code-mixed data involving Bengali-Hindi-English.  \cite{das2014identifying} present various techniques to identify languages at the token-level for the Bengali-English and Hindi-English code-mixed corpus. \cite{singh2018named} discuss various techniques to identify the named-entities in the code-mixed Hindi-English corpora consisting of 3,638 tweets. \cite{vyas2014pos} discuss various experiment to identify POS tags of the 1,062 code-mixed Hindi-English Facebook posts. They collected data from three popular celebrity Facebook public pages of Mr. Amitabh Bachchan, Mr. Shahrukh Khan, and Mr. Narendra Modi. Besides, they leverage the BBC Hindi news articles. \cite{sinha2005machine} present a rule-based machine translation system to translate the code-mixed Hindi-English sentence to monolingual Hindi and English forms. \cite{dhar2018enabling} propose a  machine translation augmentation pipeline to use on top of the standard machine translation systems. They also create a parallel corpus of 6,096 English-Hindi code-mixed sentences and their corresponding translation in English. 

In this paper, we propose a large-scale parallel corpus for code-mixed English-Hindi social media text messages. In contrast to similar works (\cite{dhar2018enabling}), the proposed dataset is significantly more extensive and comprises of multiple social media platforms. Also, the dataset spans diverse topics such as sports, entertainment, news, etc. We also showcase the limitations of state-of-the-art machine translation systems on code-mixed datasets and present a translation architecture that outperforms state-of-the-art systems.

\section{Dataset}
We use multiple code-mixing datasets (\cite{singh2018named}, \cite{swami2018corpus}, \cite{prabhu2016subword}, \cite{barman-etal-2014-code}, and \cite{senti}). One major advantage of using these datasets is the availability of high-quality code-mixed sentences without manual filtering. Also, it offers diversity in terms of the source of the data collection as most of the major social networking platforms (e.g., Twitter, Facebook, etc.) are present. It also reduces the bias towards the topics in discussion.
    
\subsection{Description, Collection, and Pre-processing}
We collect a total of 49,602 sentences from multiple sources. We then shuffle, pre-process and share these sentences with the annotators to provide the corresponding English translation. Each sentence in the corpus is written in the Roman script. Pre-processing of the dataset involves the following steps:
\begin{itemize}
    \item We remove sentences with less than five or more than 40 tokens. We introduce the upper limit on the sentence length to speed up the annotation process.
    \item We remove sentences having a percentage of out of vocabulary (OOV) words less than 50\% or more than 90\%. Lower limit (i.e., 50\%) helps to filter out the sentences with the majority of English words whereas the upper limit (i.e., 90\%) filter out the sentences with high Hindi words. We consider alphanumeric tokens as part of the vocabulary. We are using the English dictionary of the Natural Language Toolkit (NLTK) to identify the out of vocabulary words.
\end{itemize}
Post pre-processing, we obtain a total of 25,346 code-mixed sentences.

\subsection{Annotation}
The objective of the annotation is to provide the English translation of the code-mixed English-Hindi sentence. We employ 54 student annotators studying in several disciplines (Computer Science and Engineering, Electrical Engineering, etc.) at Indian Institute of Technology Gandhinagar in the annotation task. Each annotator has expert level proficiency in writing, speaking, and understanding English and Hindi languages. We assign randomly selected 400 samples to each annotator, and the annotator has to provide the corresponding translation of each sentence in English. Each sentence in the final dataset is annotated by a single annotator.  We provide a set of guidelines for each annotator for the annotation task. The annotation guidelines are listed below:
\begin{itemize}
    \item Special characters and emoticons: Use the best understanding to include or skip these symbols and characters in the translated English sentences.
    \item Links, mentions(e.g., @some\_user) and hashtags: Keep the same links, mentions, and hashtags in the translated sentence.
    \item Incorrect spellings (u, hm, pls, coz, etc.): Translated sentence should have the correct spelling for each word.
    \item Lower case: Write the translated sentence in lowercase.
    \item Proper English sentence: If the input sentence is already in English and also grammatically correct with no spelling mistakes, then its translation will only be ``\&'' (without quotes). E.g., ``I can translate the sentence quickly'', do not require any modification.
    \item Ambiguous sentence: Do not translate an ambiguous sentence. If the sentence is unclear to translate in English, mark it as ``\#'' (without quotes).
    \item Abusive words: Do not translate sentences containing abusive/cuss words. Mark it as ``\#'' (without quotes).
\end{itemize}
We refrain from translating sentences containing abusive words. Also, the annotators do not provide any translation for the ambiguous sentences. Post annotation, we obtain 21,597 sentences. It also includes sentences that are refrained from the translation (i.e., proper English sentence, ambiguous sentence, and sentences containing abusive words). Finally, we obtain 13,738 code-mixed sentences\footnote{https://doi.org/10.5281/zenodo.3605597} with the corresponding English translation. 

Figure \ref{fig:example} shows three example of the code-mixed sentence and their corresponding translation in the corpus. First two examples are the high-quality translation by the annotators and do not require any changes, whereas the translation in the last example is of poor quality as they do not satisfy the semantic requirements and require modification for the correct translation. We are not making any changes to the poor quality translation of the code-mixed sentences.

Figure \ref{fig:example_no_trans} shows three examples of the sentences that come under the refrain category of sentences for translation. Example I is a sentence that is already in English and requires no translation. The sentence in example II contains the abusive word, whereas the sentence in example III is ambiguous to translate.
\begin{figure}[!tbh]
    \centering
    \begin{tcolorbox}[colback=white]

\begin{center} 
\hl{\textbf{Example I} }\\
\end{center}
\textsc{Code-mixed Sentence}: \textcolor{alizarin}{is seat me girne ka koi chance nhi hai}\\
\textsc{English Translation}: \textcolor{cadmiumgreen}{there is no chance of falling down from this seat}\\
\textsc{Require changes in the English translation?}: \textcolor{cadmiumgreen}{No}

\begin{center} 
\hl{\textbf{Example II} }\\
\end{center}
\textsc{Code-mixed Sentence}: \textcolor{alizarin}{Thnks buds! Kabhi kabhi aajate hai achhe photos}\\
\textsc{English Translation}: \textcolor{cadmiumgreen}{Thank you buddy, sometime good photos are captured.}\\
\textsc{Require changes in the English translation?}: \textcolor{cadmiumgreen}{No}

\begin{center} 
\hl{\textbf{Example III} }\\
\end{center}
\textsc{Code-mixed Sentence}: \textcolor{alizarin}{Australia ke saath abhi jeete nahi hai, magar NZ ke saath final kaise jeetenge iss soch mein bhartiya yuvak on twitter.}\\
\textsc{English Translation}: \textcolor{cadmiumgreen}{Indian youth on twitter thinking that - We have not won against Australia yet, but how would we win final with NZ?}\\
\textsc{Require changes in the English translation?}: \textcolor{cadmiumgreen}{Yes}

\end{tcolorbox}
\caption{Example translation of the code-mixed sentences in the corpus. The annotators provide translations to the code-mixed sentences. A change in the translation is required if the translation does not meet the semantic requirements.}
    \label{fig:example}
\end{figure}

\begin{figure}[!tbh]
    \centering
    \begin{tcolorbox}[colback=white]

\begin{center} 
\hl{\textbf{Example I} }\\
\end{center}
\textsc{Code-mixed Sentence}: \textcolor{alizarin}{RT: Today is the birth anniversary of Maharana Pratap, whose bravery \& indomitable spirit doesn't fail to inspire even today.}\\
\textsc{Label}: \textcolor{cadmiumgreen}{\&}\\
\textsc{Reason for no translation}: \textcolor{cadmiumgreen}{Sentence already in English}

\begin{center} 
\hl{\textbf{Example II} }\\
\end{center}
\textsc{Code-mixed Sentence}: \textcolor{alizarin}{sach bolu ? Aap Cuss hai}\\
\textsc{Label}: \textcolor{cadmiumgreen}{\#}\\
\textsc{Reason for no translation}: \textcolor{cadmiumgreen}{Presence of abusive/cuss word in sentence.}

\begin{center} 
\hl{\textbf{Example III} }\\
\end{center}
\textsc{Code-mixed Sentence}: \textcolor{alizarin}{yuhi kat jaayega safar sath tweetne se , ki manzil aayegi nazar sath tweetne se . Hum raahi Twitter ke}\\
\textsc{Label}: \textcolor{cadmiumgreen}{\#}\\
\textsc{Reason for no translation}: \textcolor{cadmiumgreen}{Unclear/ambiguous sentence.}

\end{tcolorbox}
\caption{Example of the code-mixed sentences with no translation by the annotators.  We replace the cuss word in Example II with the word ``Cuss''.}
    \label{fig:example_no_trans}
\end{figure}

\subsection{Exploratory Analysis}
\begin{figure*}[h]
    \begin{tabular}{@{}c@{}c@{}c@{}}
      \includegraphics[width=.5\hsize]{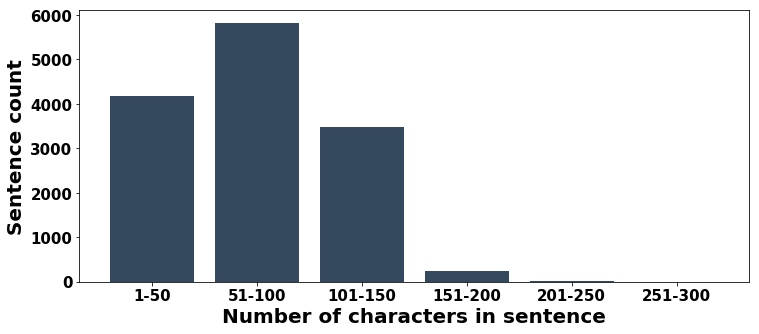} &
      \includegraphics[width=.5\hsize]{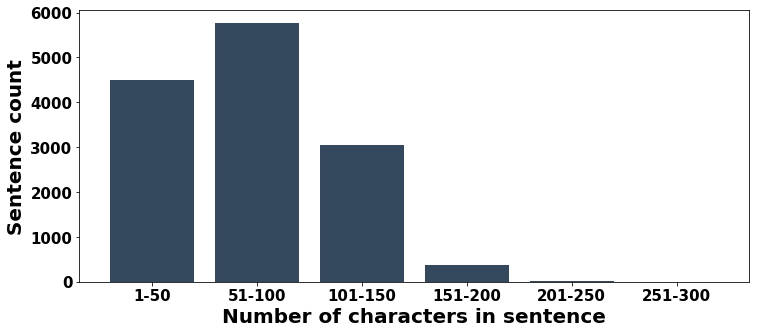} & \\
      (a) & (b)\\
    \end{tabular}
    \caption{ Distribution of the message length for the (a) code-mixed and (b) translated sentences. }\label{fig:message}
    \end{figure*}

\begin{itemize}
    
    \item \textbf{Out of vocabulary (OOV) words}: Figure \ref{fig:oov} shows the distribution of the OOV words in the code-mixed sentences. We are using the NLTK English dictionary for this study. Apart from the Romanized Hindi words, hashtags and mentions also fall into the category of OOV words. We consider alphanumeric tokens as part of the vocabulary. We have sentences with the percentage of OOV words greater than 50\% and less than 90\%. This distribution is also indicative of the non-standard writing style of the users on social media sites such as Twitter, Facebook, etc.
    
    \begin{figure}[h]
    \centering
    \includegraphics[width=1\linewidth]{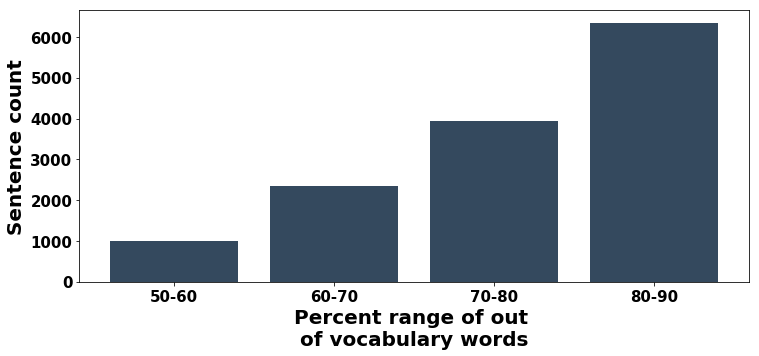}
    \caption{Distribution of out of vocabulary words in the code-mixed messages.}
    \label{fig:oov}
    \end{figure}

    \item \textbf{Code Mixing Index (CMI)}: CMI is the metric introduced by \cite{das2014identifying}. It is the measure of the degree of code-mixing in a corpus. CMI is calculated as follows
    
    \[ CMI= \begin{cases} 
          100 * [1- \frac{max(w_{i})}{n-u}] & n> u \\
          0 & n=u 
       \end{cases}
    \]
    
    Here, $w_{i}$ is the number of words of the language $i$, max\{{$w_{i}$}\} represents the number of words of the most prominent language, $n$ is the total number of tokens, $u$ represents the number of language-independent tokens (such as named entities, abbreviations, mentions, hastags, etc.). CMI value range from 0 to 100. A value close to 0 suggest multilingualism in the corpus, whereas high CMI values indicate a high degree of code-mixing. To calculate the value of CMI, we randomly sample 50 code-mixed sentences from the corpus and annotate them at the token level with three language tags English, Hindi, and Other. The CMI calculated for this set of sentences is 77.

    \item \textbf{Topics}: Figure \ref{fig:wc} shows the word cloud of the code-mixed and English translated sentences. It is evident from the word cloud that words from multiple domains such as politics, entertainment, sports, etc., are very frequently used. We remove the hyperlinks, mentions, and hashtags from the translated sentences to create the word cloud.

   \begin{figure}[!htbp]
    \centering
    \begin{tabular}{@{}c@{}c@{}c@{}}
      \includegraphics[width=0.5\linewidth]{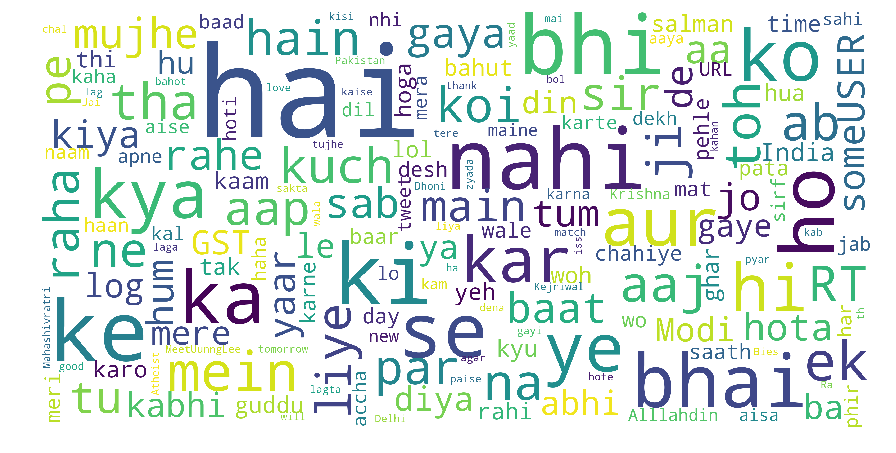} & 
      \includegraphics[width=0.5\linewidth]{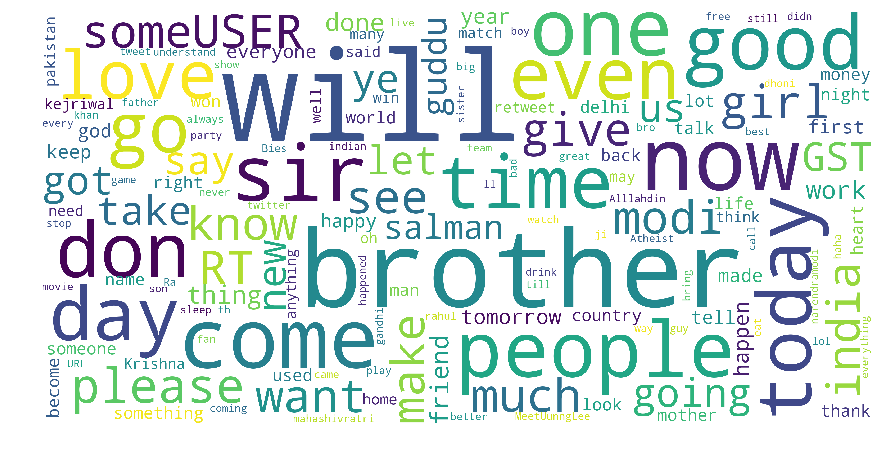} &\\
      (a) & (b) \\
    \end{tabular}
    \caption{Word cloud of the (a) code-mixed and (b) translated sentences.}\label{fig:wc}
    \end{figure}

    \item \textbf{Message length}: Figure \ref{fig:message} shows the distribution of the message length for the code-mixed and the translated sentences. Distribution of message length for code-mixed and the translated sentences follows a similar trend.

    \item \textbf{Quality of Translations (QT)}: To evaluate the quality of the translations by the annotators, we have randomly sampled 50 sentences from the corpus. We provide two labels to each of the translation \textit{correct translation} and \textit{require change}. The quality of translation is calculated as follows
    $${QT}=\frac{Count \: of \: correct\:  translations}{Sample \: size}$$
    42 samples out of 50 do not require any changes. Thus, the quality of translation is \textit{0.84}.
\end{itemize}

\section{Evaluation of Machine Translation Systems}
We experiment with various machine translation systems and evaluate their performance on our proposed corpus. The majority of these systems perform well for the monolingual translation tasks. However, these systems demonstrate severe limitations in translating code-mixed text prevalent on social media platforms. In the following experiments, we randomly sample 50 code-mixed sentences from the corpus. We use human translated sentences as reference.
    
\begin{itemize}
    \item \textbf{Bing Translate (BT)}: We evaluate the performance of BT on the code-mixed corpus. We set the language of the code-mixed input sentence as Hindi on the BT platform. The BLEU-1 score for this set of sentences is 0.139. This value is close to 0 and suggests the poor performance of the system on the code-mixed data.
    \item \textbf{Google Translate (GT)}:  Next, we evaluate the performance of the GT on the code-mixed corpus. We set GT to auto-detect the language of the code-mixed input sentence. The BLEU-1 score for this set of sentences is 0.14. Even though we observe a slightly higher value of BLEU-1 for GT, the translations can still be improved.
    \item \textbf{Proposed Pipeline + Google Translate (PPGT)}: We propose a pipeline to use on top of GT. This pipeline helps to improve the quality of input being fed to GT. Various steps of PPGT translation are:
    \begin{itemize}
        \item We provide a label for each token of the code-mixed sentence based on the language ( \textit{English, Hindi, and Other}).
        \item We create chunks of Type-I using Hindi tokens with at most two English/Other token allowed to be part of any chunk. A chunk of Type-I starts with a Hindi token.
        \item We create chunks of Type-II using the tokens that are labeled as English/Others and not part of any Type-I chunk. 
        \item We only translate the Type-I chunks using GT. We keep the chunks of Type-II as it is.
    \end{itemize}
    
Figure \ref{fig:example_pipeline} shows the example translation of code-mixed sentences using all the three techniques BT, GT, and PPGT. The Type-I and Type-II chunks in each example are used for translating the sentence using the PPGT technique. In PPGT, we maintain the original order of the chunks as that of the code-mixed sentence while translating. \newline For instance, the order of the chunks for Example II in Figure \ref{fig:example_pipeline} is \textit{[[par if its], [possible and any other guest needs a room ,], [mera room de de kisi ko bhi]]}.
\newline The BLEU-1 score of this system is 0.153, which is an improvement over both the other systems.
\end{itemize}

As most of the state-of-the-art machine translation systems do not perform well on the code-mixed data, we can build pipelines on top of these systems that can preprocess the input to these systems. \cite{dhar2018enabling} also present one such augmentation pipeline to improve the performance of these systems. These pipelines can address the challenges to code-mixed machine translation, as outlined in the next section.

\section{Challenges to Code-Mixed Machine Translation}
The nature of code-mixed text presents several challenges to the various natural language processing techniques. Google translate, one of the most sophisticated systems for translating texts in one language to another, also fails at times to translate code-mixed texts efficiently. Figure \ref{fig:translate} shows the three different instances of translating text using Google translate. Example I and II are relatively better translations as compared to example III.  We use the auto-detection of language by Google translate in all the translations. Table ~\ref{BLEU} shows the comparison of the BLEU score of all the three examples in Figure \ref{fig:translate}. The value of the BLEU score can vary between 0 and 1. Some prominent reasons for the failure of the standard machine translations systems on the code-mixed data are:
\begin{itemize}
    \item\textbf{Ambiguity in language identification}: Hindi words written in the Roman script present some significant challenges to identify the language of the text at the token level.
    \newline \textit{is, me, to} are some examples of the words that are ambiguous to classify as English and Hindi without proper knowledge of context. \newline Hashtags are often used on the social medium platforms, and code-mixed hashtags make it challenging to identify the boundaries of code-switching.
    
    \item \textbf{Spelling variations}: Romanized Hindi also presents a challenge with no standard spelling of the words. Various spellings for the same word is used based on the user's pronunciation of the word, emotions, etc. \newline E.g., \textit{jaldi, jldi, jldiii,..} are some variations for the word \textit{hurry} in English.\newline At times, people use repeated instances of some particular character to emphasize emotion, such as in \textit{jaldiii}.
    
    \item \textbf{Informal style of writing}: Writing style of the users on social media sites is informal. At times, we do not follow the standard rules of sentence structure on these platforms. This presents a challenge to translate the sentence in monolingual style where the formal sentence structure is required for the semantic purpose. \newline E.g., \textit{Sad kabhi dekha h usko.. me never}. \newline The corresponding English translation is \textit{Have you ever seen him sad? I have never seen him sad}.
    
    \item \textbf{Misplaced/ skipped punctuation}: In the informal writing style on social media platforms, punctuations are usually skipped, misplaced, or repeatedly used to express an opinion, and that makes it difficult for the machine translation system to translate such sentences. \newline E.g., \textit{Aap kb se cricket khelne lage..never saw u bfr}. \newline The sentence in the example misses a question mark(?) apart from other necessary modifications to make the structure of the sentence correct. 
    
    \item \textbf{Missing context}: Lack of knowledge of the context makes the task of machine translation difficult and challenging. Hidden sarcasm might get unnoticed while translating the sentence with missing context. \newline E.g., \textit{Note kr lijiye.. Bandi chal rahi h ;)} is a code-mixed sentence and demonetisation (notebandi) is the hidden context.

\end{itemize}

\section{FAIR Principles}
We are making the dataset FAIR by employing the various guiding principles\footnote{https://www.force11.org/fairprinciples}. 
\begin{itemize}
    \item \textbf{Findable}: A globally unique and eternally persistent identifier for our data is available for access\footnote{https://doi.org/10.5281/zenodo.3605597}. We describe our data with rich metadata, and the previous section explains all the exploratory analysis for the metadata. The data is indexed in OpenAIRE\footnote{http://bit.ly/2FH4hDU}.
    \item \textbf{Accessible}: The metadata and the data are retrievable and accessible by humans and machines by the identifier using a standardized communications protocol such as HTTP.  
    \item \textbf{Interoperable}: The metadata and the data use the English language for knowledge representation, which is formal, accessible, shared, and broadly applicable language. 
    \item \textbf{Re-usable}: The data is released with a clear and accessible data usage license\footnote{https://creativecommons.org/licenses/by/4.0/legalcode}.  Metadata and the data are sufficiently well-described and rich.
\end{itemize}
\section{Conclusion}
In this paper, we present a parallel corpus for the English-Hindi code-mixed machine translation task. We discuss various challenges that state of the art machine translation system build for monolingual corpus face while dealing with code-mixed corpora. We also demonstrate the performance of the various systems on our parallel corpus. It is evident that we have a large scope to build systems that can overcome the challenges associated with the code-mixed text from various social media platforms.
\begin{figure}[!tbh]
    \centering
    \begin{tcolorbox}[colback=white]

\begin{center} 
\hl{\textbf{Example I} }\\
\end{center}
\textsc{Code-mixed Sentence}: \textcolor{alizarin}{@Prankoholic tumko matlab kya time hai din ka, kuch samaj nahi aata na}\\
\textsc{Type-I chunks}: \textcolor{cadmiumgreen}{[tumko matlab kya time hai din ka, kuch samaj nahi aata na]}\\
\textsc{Type-II chunks}: \textcolor{cadmiumgreen}{[@Prankoholic]}\\
\textsc{English Translation using BT}: \textcolor{alizarin}{@prankoholic what time do you mean of the day, some society does not come.}\\
\textsc{English Translation using GT}: \textcolor{alizarin}{@Prankoholic you mean what is the time of day, don't understand anything}\\
\textsc{English Translation using PPGT}: \textcolor{alizarin}{@Prankoholic Do you mean what is the time of day, no sense}

\begin{center} 
\hl{\textbf{Example II} }\\
\end{center}
\textsc{Code-mixed Sentence}: \textcolor{alizarin}{par if its possible and any other guest needs a room , mera room de de kisi ko bhi}\\
\textsc{Type-I chunks}: \textcolor{cadmiumgreen}{[par], [mera room de de kisi ko bhi]}\\
\textsc{Type-II chunks}: \textcolor{cadmiumgreen}{[if its possible and any other guest needs a room ,]}\\
\textsc{English Translation using BT}: \textcolor{alizarin}{On if its possible egg any
other guest needs coming room , my room day to anyone}\\
\textsc{English Translation using GT}: \textcolor{alizarin}{par if its possible and any
other guest needs a room ,
mera room de de kisi ko bhi}\\
\textsc{English Translation using PPGT}: \textcolor{alizarin}{par if its possible and any
other guest needs a room , Give my room to anyone}

\begin{center} 
\hl{\textbf{Example III} }\\
\end{center}
\textsc{Code-mixed Sentence}: \textcolor{alizarin}{@UPGovt @yadavakhilesh 
Great progress Sir. Iss baar bhi aap kuch nahi karoge. }\\
\textsc{Type-I chunks}: \textcolor{cadmiumgreen}{[Iss baar bhi aap kuch nahi karoge.]}\\
\textsc{Type-II chunks}: \textcolor{cadmiumgreen}{[@UPGovt @yadavakhilesh 
Great progress Sir.]}\\
\textsc{English Translation using BT}: \textcolor{alizarin}{@upgovt @yadavakhilesh great progress sir. This time too you karoge nothing.}\\
\textsc{English Translation using GT}: \textcolor{alizarin}{@UPGovt @yadavakhilesh
Great progress Sir. Iss baar bhi aap kuch nahi karoge.
}\\
\textsc{English Translation using PPGT}: \textcolor{alizarin}{@UPGovt @yadavakhilesh 
Great progress Sir. You will not do anything this time.}

\end{tcolorbox}
\caption{Example translation of code-mixed sentences using BT, GT, and PPGT. We use Type-I and Type-II chunks while translating the code-mixed sentence using PPGT.}
    \label{fig:example_pipeline}
\end{figure}

\begin{figure}[!tbh]
    \centering
    \begin{tcolorbox}[colback=white]

\begin{center} 
\hl{\textbf{Example I} }\\
\end{center}
\textsc{Sentence}: \textcolor{alizarin}{Bahot time baad suna ye! Achha laga}\\
\textsc{Google Translation}: \textcolor{cadmiumgreen}{Heard this after a lot of time! It was nice}\\
\textsc{Human Translation}: \textcolor{cadmiumgreen}{Heard this after a long time! It was nice}\\
\textsc{Performance of Google translate}: \textcolor{cadmiumgreen}{Good}

\begin{center} 
\hl{\textbf{Example II} }\\
\end{center}
\textsc{Sentence}: \textcolor{alizarin}{Is shaher ko ye Hua kya hai.. Kahi rakh hai to kahi dhua dhua.. Play interrupted due to bad weather}\\
\textsc{Google Translation}: \textcolor{cadmiumgreen}{What has happened to this city .. If there is smoke somewhere, then smoke somewhere .. Play interrupted payable then bad weather}\\
\textsc{Human Translation}: \textcolor{cadmiumgreen}{What has happened to this city. there is ash and smoke everywhere. play interrupted due to bad weather}\\
\textsc{Performance of Google translate}: \textcolor{cadmiumgreen}{Average}

\begin{center} 
\hl{\textbf{Example III} }\\
\end{center}
\textsc{Sentence}: \textcolor{alizarin}{Bhai IIT wale hai pehle relationship toh bane laundon ki, break up par nacha rahe ho.}\\
\textsc{Google Translation}: \textcolor{cadmiumgreen}{Brother-in-law is the first relationship to be made of laundries, you are dancing on the brake sub.}\\
\textsc{Human Translation}: \textcolor{cadmiumgreen}{Brother, you are an IITian. First get in to a relation. Then you can worry about break up.}\\
\textsc{Performance of Google translate}: \textcolor{cadmiumgreen}{Poor}

\end{tcolorbox}
\caption{Comparison of translation of code-mixed sentences by Google translate and human annotators.}
    \label{fig:translate}
\end{figure}

\begin{table}[!tbh]
\resizebox{\hsize}{!}{
\begin{tabular}{|c|l|l|l|l|}
\hline
\multicolumn{1}{|l|}{\textbf{}} & \textbf{BLEU 1} & \textbf{BLEU 2} & \textbf{BLEU 3} & \textbf{BLEU 4} \\ \hline
\textbf{Example I}              & 0.814           & 0.704           & 0.565           & 0.388           \\ \hline
\textbf{Example II}             & 0.560           & 0.362           & 0.191           & 0.152           \\ \hline
\textbf{Example III}            & 0.272           & 0.095           & 0.050           & 0 \\ \hline
\end{tabular}
}
\caption{BLEU score for various examples of Figure \ref{fig:translate}}
\label {BLEU}
\end{table}

\section{Future Work}
We can use the parallel corpus to build efficient machine translation systems for social media platforms. We can also explore various other code-mixed languages, especially those that are low resource and endangered. We can also extend the corpus presented here for various other code-mixing tasks such as language identification, named-entity recognition, etc.   

\bibliography{PHINC}

\begin{thebibliography}{}

\bibitem[\protect\citeauthoryear{Barman \bgroup et al\mbox.\egroup
  }{2014}]{barman-etal-2014-code}
Barman, U.; Das, A.; Wagner, J.; and Foster, J.
\newblock 2014.
\newblock Code mixing: A challenge for language identification in the language
  of social media.
\newblock In {\em Proceedings of the First Workshop on Computational Approaches
  to Code Switching},  13--23.
\newblock Doha, Qatar: Association for Computational Linguistics.

\bibitem[\protect\citeauthoryear{Das and
  Gamb{\"a}ck}{2014}]{das2014identifying}
Das, A., and Gamb{\"a}ck, B.
\newblock 2014.
\newblock Identifying languages at the word level in code-mixed indian social
  media text.
\newblock In {\em Proceedings of the 11th International Conference on Natural
  Language Processing},  378--387.

\bibitem[\protect\citeauthoryear{Dhar, Kumar, and
  Shrivastava}{2018}]{dhar2018enabling}
Dhar, M.; Kumar, V.; and Shrivastava, M.
\newblock 2018.
\newblock Enabling code-mixed translation: Parallel corpus creation and mt
  augmentation approach.
\newblock In {\em Proceedings of the First Workshop on Linguistic Resources for
  Natural Language Processing},  131--140.

\bibitem[\protect\citeauthoryear{Prabhu and Verma}{2016}]{prabhu2016subword}
Prabhu, Ameya, J. A. S.~M., and Verma, V.
\newblock 2016.
\newblock Towards sub-word level compositions for sentiment analysis of
  hindi-english code mixed text.
\newblock {\em arXiv preprint arXiv:1611.00472}.

\bibitem[\protect\citeauthoryear{Singh \bgroup et al\mbox.\egroup
  }{2018}]{singh2018named}
Singh, V.; Vijay, D.; Akhtar, S.~S.; and Shrivastava, M.
\newblock 2018.
\newblock Named entity recognition for hindi-english code-mixed social media
  text.
\newblock In {\em Proceedings of the Seventh Named Entities Workshop},  27--35.

\bibitem[\protect\citeauthoryear{Sinha and Thakur}{2005}]{sinha2005machine}
Sinha, R. M.~K., and Thakur, A.
\newblock 2005.
\newblock Machine translation of bi-lingual hindi-english (hinglish) text.
\newblock {\em 10th Machine Translation summit (MT Summit X), Phuket, Thailand}
   149--156.

\bibitem[\protect\citeauthoryear{Swami \bgroup et al\mbox.\egroup
  }{2018}]{swami2018corpus}
Swami, S.; Khandelwal, A.; Singh, V.; Akhtar, S.~S.; and Shrivastava, M.
\newblock 2018.
\newblock A corpus of english-hindi code-mixed tweets for sarcasm detection.
\newblock {\em arXiv preprint arXiv:1805.11869}.

\bibitem[\protect\citeauthoryear{Vrishank~Shete and Mittal}{2016}]{senti}
Vrishank~Shete, G. S.~C., and Mittal, L.
\newblock 2016.
\newblock Sentiment analysis on hindi-english code mixed data using svm.
\newblock [Online; accessed 08-Jan-2020].

\bibitem[\protect\citeauthoryear{Vyas \bgroup et al\mbox.\egroup
  }{2014}]{vyas2014pos}
Vyas, Y.; Gella, S.; Sharma, J.; Bali, K.; and Choudhury, M.
\newblock 2014.
\newblock Pos tagging of english-hindi code-mixed social media content.
\newblock In {\em Proceedings of the 2014 Conference on Empirical Methods in
  Natural Language Processing (EMNLP)},  974--979.

\end{thebibliography}
\bibliographystyle{aaai}

\end{document}